\documentclass[11pt]{article}

\usepackage[preprint]{acl}

\usepackage{times}
\usepackage{latexsym}

\usepackage[T1]{fontenc}

\usepackage[utf8]{inputenc}

\usepackage{microtype}

\usepackage{inconsolata}

\usepackage{graphicx}
\usepackage{amsmath}
\usepackage{amsfonts}
\usepackage{pifont}
\numberwithin{equation}{section}
\usepackage{bbm}
\usepackage{booktabs}
\usepackage{multirow}
\usepackage{colortbl}
\usepackage{subcaption}

\usepackage{listings}
\usepackage{xcolor}
\usepackage{caption}

\newcommand{\Tref}[1]{Tab.~\ref{#1}}
\newcommand{\Eref}[1]{Eq.~(\ref{#1})}
\newcommand{\Fref}[1]{Fig.~\ref{#1}}
\newcommand{\Sref}[1]{Sec.~\ref{#1}}

\newcommand{\ie}{\textit{i}.\textit{e}.}
\newcommand{\eg}{\textit{e}.\textit{g}.}

\newcommand{\toolns}{\textsc{GuardAD}}
\newcommand{\tool}{\toolns\space}
\newcommand{\admlmns}{AD-MLLMs}
\newcommand{\admlm}{\admlmns\space}

\newcommand{\admlmsinglens}{AD-MLLM}
\newcommand{\admlmsingle}{\admlmsinglens\space}

\title{\toolns: Safeguarding Autonomous Driving MLLMs via \\Markovian Safety Logic}

\author{
  Tianyuan Zhang$^\text{1}$, Peng Yue$^\text{1}$, Zihao Peng$^\text{1}$, Jiangfan Liu$^\text{1}$, Zonghao Ying$^\text{1}$, Jiakai Wang$^\text{2}$, \\\textbf{Tianlin Li$^\text{1}$, Jian Yang$^\text{1}$, Yaodong Yang$^\text{3}$, Aishan Liu$^{\text{1},*}$, Xianglong Liu$^{\text{1,2}}$} \\
  $^\text{1}$Beihang University  $^\text{2}$Zhongguancun Laboratory  $^\text{3}$Peking University\\
  $^\text{*}$Correponding Author\\
  \texttt{zhangtianyuan@buaa.edu.cn} \\
  }

\begin{document}
\maketitle
\begin{abstract}
Multimodal large language models (MLLMs) are increasingly integrated into autonomous driving (AD) systems; however, they remain vulnerable to diverse safety threats, particularly in accident-prone scenarios. Recent safeguard mechanisms have shown promise by incorporating logical constraints, yet most rely on static formulations that lack temporally grounded safety reasoning over evolving traffic interactions, resulting in limited robustness in dynamic driving environments. To address these limitations, we propose \toolns, a model-agnostic safeguard that formulates AD safety as an evolving Markovian logical state. \tool introduces Neuro-Symbolic Logic Formalization, which represents safety predicates over heterogeneous traffic participants and continuously induces them via $n$-th order Markovian Logic Induction. This design enables the inference of emerging and latent hazards beyond single-step observations. Rather than simply vetoing unsafe actions, \tool performs Logic-Driven Action Revision, where inferred safety states actively guide action refinement without modifying the underlying MLLM. Extensive experiments on multiple benchmarks and \admlm demonstrate that \tool substantially reduces accident rates (-32.07\%) while slightly improving task performance (+6.85\%). Moreover, closed-loop simulation evaluations, together with physical-world vehicle studies, further validate the effectiveness and potential of \toolns.

\end{abstract}

\section{Introduction}
\label{sec:introduction}

Driven by their strong generalization ability and improved interpretability, Multi-modal Large Language Models (MLLMs) have recently been adopted in a wide range of domains, including autonomous driving (AD) \cite{sima2024drivelm, ma2024dolphins}. AD-specific MLLMs (\admlmns) can handle complex driving tasks that span the perception-prediction-planning pipeline, such as scene understanding, behavior forecasting, and trajectory generation, thereby providing a new paradigm for building end-to-end AD systems.

Despite this, \admlm are facing diverse safety threats in realistic deployments \cite{ying2026safebench}. For example, accident-prone scenarios can substantially degrade \admlmsinglens's performance, leading to unsafe accidents \cite{kim2025vru}. Existing logic-driven safeguards for \admlm have shown promise in enforcing safe and stable behavior under diverse threats \cite{zhangsafeauto,liang2026t2vshield}. Nevertheless, most existing safeguards only rely on static and pre-defined constraints (\eg, ``stop when the light is red'') and struggle to adapt to dynamic and complex driving environments. The key problem is that such failures often stem not from isolated perception errors, but from the lack of temporally grounded safety reasoning over evolving traffic interactions. 

\begin{figure}
    \centering
    \includegraphics[width=0.98\linewidth]{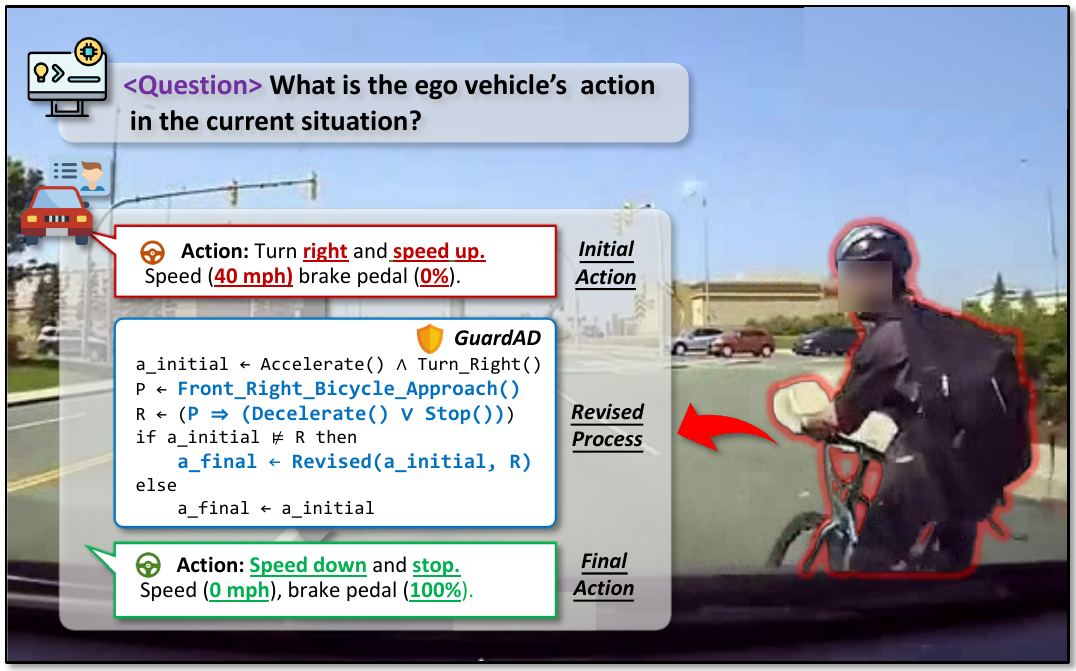}
    \caption{Illustration of \toolns, a safeguard that captures the dynamic nature of autonomous driving and mitigates accident-prone decisions.}
    \label{fig:frontpage}
    \vspace{-0.15in}
\end{figure}


To address these, we introduce \toolns, a model-agnostic safeguard that formulates AD safety as an evolving Markovian logical state. Specifically, we conceptualize AD safety not as a set of static, instantaneous constraints, but as a latent logical state that evolves with the interaction history between the ego vehicle and its environment. Based on this view, \tool defines a Neuro-Symbolic Logic Formalization that instantiates safety predicates over heterogeneous traffic participants and their spatial relations, and maintains a structured safety state at each decision step. This state is updated via an $n$-th order Markovian Logic Induction by aggregating multi-step predicate evidence, enabling \tool to infer emerging hazards that are not observable from the current frame alone. Rather than directly vetoing unsafe actions, \tool performs Logic-Driven Action Revision, where the inferred safety state guides the construction of alternative actions while remaining fully decoupled from the underlying \admlmns.
We evaluate \tool on DriveLM \cite{sima2024drivelm} and VRU-Accident \cite{kim2025vru} benchmarks. Across models, \tool substantially reduces accident rate (-32.07\%) while slightly improving task performance (+6.85\%). Closed-loop evaluations in CARLA further confirm these gains, showing a pronounced improvement in success rate (+45.60\%). Moreover, our human-in-the-loop evaluation with a physical-world vehicle shows that \tool narrows the performance gap between \admlm and human drivers. Our \textbf{contributions} are summarized as follows:

\begin{itemize}
    \vspace{-0.10in}
    \item We propose \toolns, a model-agnostic safeguard that formulates AD safety as an evolving Markovian logical state.
    \vspace{-0.12in}
    \item We develop a Neuro-Symbolic Logic Formalization that encodes safety states and update them via an $n$-th order Markovian Logic Induction, enabling structured constraint inference and Logic-driven Action Revision.
    \vspace{-0.12in}
    \item Extensive experiments on multiple benchmarks and \admlm show that \tool reduces accident rates while preserving task performance and usability.
\end{itemize}
\section{Backgrounds and Preliminaries}
\label{sec:backgrounds}

\quad \textbf{\admlmns.}
Recent work applies \admlm for planning, decision-making, \textit{etc}. Representative efforts include end-to-end control and multimodal reasoning with DriveGPT4/DriveGPT4-V2 \cite{xu2024drivegpt4,xu2025drivegpt4v2}, knowledge- and CoT-enhanced driving with DriveMLM and Dolphins \cite{wang2023drivemlm,ma2024dolphins}, and hierarchical planning with DriveVLM \cite{tian2024drivevlm}. Senna and DriveLM incorporate QA into formulations \cite{jiang2024senna,sima2024drivelm}, ORION studies closed-loop VQA generation \cite{fu2025orion}, and other lines explore richer representations and efficient deployment \cite{wang2025omnidrive,xie2025s4driver,gopalkrishnanmulti,hegde2025distilling}.

At each time step $t$, the \admlmsingle receives a short history of multi-modal inputs
$\mathbb{O}_{t-k:t} = \bigcup_{\tau=t-k}^{t} \{(\mathbf{I}_\tau,\, \mathbf{T}_\tau)\}$,
where $\mathbf{I}_t$ denotes the image and $\mathbf{T}_t$ denotes the corresponding textual query or instruction (\eg, route description). Let $\mathbb{A}$ denote the action space. Conditioned on $\mathbb{O}_{t-k:t}$, the base \admlmsingle outputs an action distribution $\pi(\mathbf{a}_t \mid \mathbb{O}_{t-k:t})$, and the decision is obtained by
\begin{equation}
\label{equ:action}
\mathbf{a}_t^{\text{base}}=\arg\max_{\mathbf{a}\in\mathbb{A}} \pi(\mathbf{a}\mid \mathbb{O}_{t-k:t}).
\end{equation}

\noindent \textbf{Problem Definition.}
We study safeguards for \admlm under accident-prone external threats. Under such threats, the base \admlmsingle may output an unsafe action $\mathbf{a}_t^{\text{unsafe}}\in\mathbb{A}$ from multi-modal observations $\mathbb{O}_{t-k:t}$. 
Our goal is to design a model-agnostic safeguard that post-processes $\mathbf{a}_t^{\text{unsafe}}$ into a final action $\mathbf{a}^{safe}_t\in\mathbb{A}$, reducing safety-critical failures while preserving task performance.

\section{Related Work}
\label{sec:related_work}

AD-relevant models face significant security issues \cite{zhang2024lanevil, zhang2024towards, liu2025metadv, zhang2025towards, liu2026adversarial, liu2025natural, kong2025universal}. 
Safety enhancements for AD-relevant models include adversarial training \cite{zhang2023cat, zhang2024module,lu2025adversarial}, fine-tuning under perturbations \cite{liao2025robodrivevlm, oh2024towards}, and simulation testing \cite{cai2025text2scenario, liu2025adversarial}. While effective, they mainly target perception-level errors \cite{liang2024object} rather than grounded safety reasoning over evolving traffic interactions.
Logic-driven safeguards have been studied in embodied agents \cite{yin2025safeagentbench, xiang2024guardagent, wang2025agentspec, xiang2025guardagent, qin2025robofactory, zhou2025code, liu2024compromising, liu2025agentsafe,liang2025safemobile}, but they are largely designed for specific tasks.

For \admlmns, SafeAuto \cite{zhangsafeauto} is the closest prior work, using visual structured knowledge and fixed traffic rules to correct high-level decisions. However, SafeAuto is ego- and rule-centric, largely static, and does not model temporally evolving risks; moreover, it is tightly coupled to a specific backbone and interface, limiting transferability across \admlmns. As a result, general safeguards for \admlm in dynamic, multi-agent driving remain underexplored.


\textbf{Comparisons.}
\tool differs from prior works by targeting the dynamic, interactive nature of AD instead of enforcing static traffic rules, and by incorporating temporal context beyond single-step observations. It is also model-agnostic rather than a specialized, model-coupled design.
\section{Method}
\label{sec:method}

\begin{figure*}[htp]
    \centering
    \includegraphics[width=0.98\linewidth]{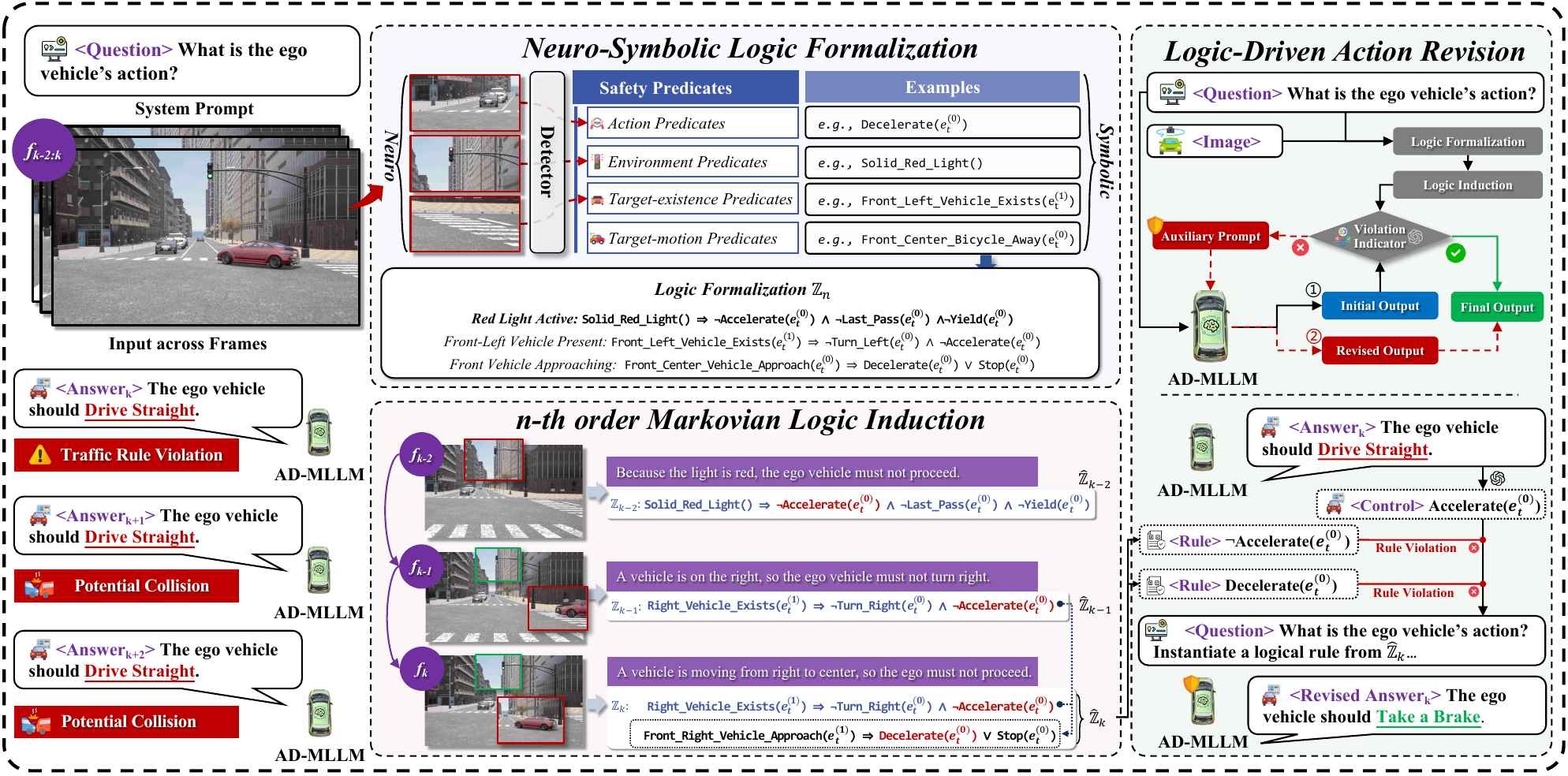}
    \vspace{-0.05in}
    \caption{Framework of \toolns. \tool is a model-agnostic safeguard that formulates AD safety as an evolving Markovian logical state. It constructs the safety state via Neuro-Symbolic Logic Formalization, induces a temporally consistent state through $n$-th order Markovian Logic Induction, and then performs Logic-Driven Action Revision to correct unsafe actions without modifying the \admlmns.}
    \vspace{-0.1in}
    \label{fig:framework}
\end{figure*}

\quad \textbf{Motivation.}
Recent methods \cite{zhangsafeauto} mitigate risky actions, but they are often based on static rules and fail to capture the dynamics of driving environments.
Motivated by these, we proposed \tool (as shown in \Fref{fig:framework}), conceptualize AD safety not as a set of static, instantaneous constraints, but as a logical state that evolves with the interaction history between the ego vehicle and environment. \tool is a model-agnostic safeguard that formulates AD safety as an evolving Markovian logical state. It encodes Markovian Safety Logic and updates the state with an $n$-th order Markovian Logic Induction, then performs Logic-driven Action Revision to refine unsafe actions without modifying the \admlmns.

\subsection{Neuro-Symbolic Logic Formalization}
\label{sec:neuro_logic_formalization}

To enable safety reasoning beyond perception-level, \tool lifts neural perception and policy signals into a structured symbolic interface. Concretely, given the front-view image $\mathbf{I}_t$, an neural perception stack produces a set of grounded entities
$\mathbb{E}_t = \{e_{t}^{(0)}, e_{t}^{(1)}, \ldots, e_{t}^{(N)}\}$,
where $e_{t}^{(0)}$ denotes the ego vehicle and the remaining entities correspond to surrounding traffic participants (\eg, vehicles, pedestrians, and cyclists). For each entity $e_{t}^{(i)}$, we obtain geometric and semantic attributes from the neural outputs, such as relative position, heading/motion direction, and motion cues. In parallel, the base \admlmsingle provides the ego action hypothesis used to instantiate action-related predicates.

On top of these attributes, \tool constructs a finite set of safety predicates $\mathbb{P}=\{\mathcal{P}_1,\ldots,\mathcal{P}_M\}$, where each predicate is a discrete, interpretable indicator that evaluates safety-relevant properties (\ie, boolean-valued tests over entities and context). The predicates cover four common categories, each capturing a different safety-relevant aspect of the scene. Specifically, the \ding{182} \textit{Action Predicates} (\eg, \texttt{Decelerate}) describe the current or intended ego driving status implied by the \admlm output; the \ding{183} \textit{Environment Predicates} (\eg, \texttt{Solid\_Red\_Light}) encode environmental constraints such as traffic signs and traffic lights; the \ding{184} \textit{Target-existence Predicates} (\eg, \texttt{Front\_Left\_Region\_Vehicle\_Exists}) indicate the presence of other participants in predefined regions; and the \ding{185} \textit{Target-motion Predicates} (\eg, \texttt{Front\_Center\_Bicycle\_Away}) characterize the movement trends of other participants.

We denote by $\mathbb{Z}_t$ the instantaneous safety logical state at time $t$, which summarizes the activated safety constraints grounded from the current observation. \tool first determines the visually instantiated entity subset $\mathbb{E}_t^{\text{vis}} \subseteq \mathbb{E}_t$, and then evaluates a context-relevant predicate subset $\mathbb{P}_t^{(i)} \subseteq \mathbb{P}$ for each $e_t^{(i)} \in \mathbb{E}_t^{\text{vis}}$.
Safety constraints are activated through Horn-style rule templates of the form $\bigwedge_{\mathcal{P}\in\mathbb{S}^j}\mathcal{P}(e_t^{(i)}) \Rightarrow z$,
where $\mathbb{S}^{(j)}$ is a predicate conjunction and $z$ is a discrete safety constraint. The resulting state $\mathbb{Z}_t$ can be written compactly as
\begin{equation}
\label{equ:define_z}
\big\{
    z \big|
    \exists\, e_{t}^{(i)} \in \mathbb{E}_t^{\mathrm{vis}},
    \exists\, \mathbb{S}^{(j)} \subseteq \mathbb{P}_t^{(i)},
    \wedge_{\mathcal{P}\in\mathbb{S}^{(j)}}\mathcal{P}(e_{t}^{(i)}) \Rightarrow z
\big\},
\end{equation}
where ``$\Rightarrow$'' denotes rule instantiation rather than a truth-functional implication: whenever the predicate conjunction in the antecedent holds for some grounded entity, the corresponding constraint $z$ is activated and included in $\mathbb{Z}_t$.
This formulation makes explicit how the neural-to-symbolic lifting deterministically compiles the current observation into active safety constraints, yielding a symbolic state that can be consumed by subsequent temporal induction and action revision.

\begin{figure}[t]
    \centering
    \includegraphics[width=0.98\linewidth]{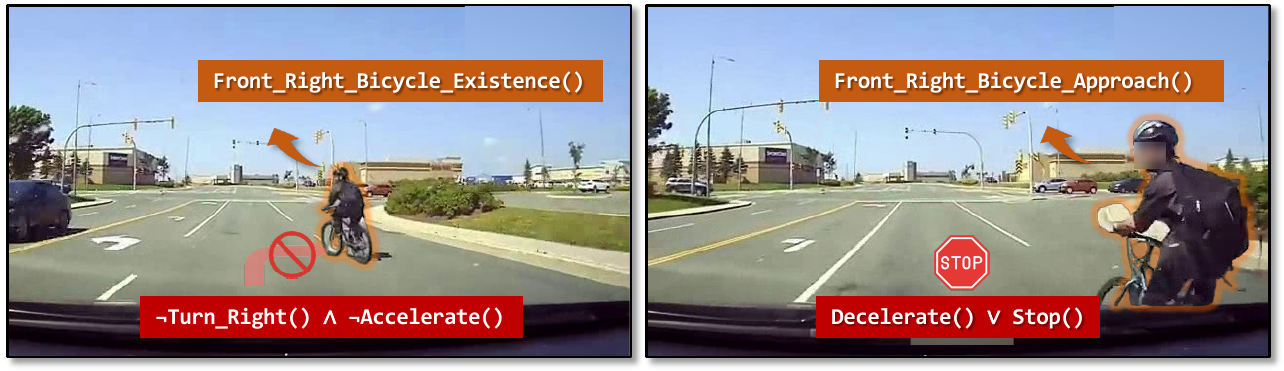}
    \caption{Example of neuro-symbolic logic formalization. An approaching cyclist is lifted to the predicate \texttt{Front\_Center\_Bicycle\_Approaching}.}
    \vspace{-0.15in}
    \label{fig:example}
\end{figure}

\textbf{Example.} As shown in \Fref{fig:example}, a safety predicate constraint such as ``if a cyclist is approaching in front, then the ego vehicle should stop or decelerate'' can be expressed as the logical rule implication
\texttt{Front\_Center\_Bicycle\_Approach}($e_{t}^{\text{bic}}$) $\Rightarrow$ \big(\texttt{Decelerate}($e_{t}^{(0)}$) $\lor$ \texttt{Stop}($e_{t}^{(0)}$)\big), where $e_{t}^{(0)}$ denotes the ego vehicle, $e_{t}^{\text{bic}}$ denotes a cyclist detected in the front-center region at time $t$. These derived predicates are constraints within $\mathbb{Z}_t$.
\textit{More details of the safety constraint are in the Appendix.}

\subsection{\textit{n}-th order Markovian Logic Induction}
\label{sec:markov}

While $\mathbb{Z}_t$ captures the safety logical states activated by predicates in the current frame, many risk-inducing patterns in AD (\eg, approaching vehicles or crossing pedestrians) evolve over time. To incorporate such temporal structure, \tool uses an $n$-th order Markov Logic Network (nMLN) to refine the instantaneous state into a temporally induced one, which we denote by $\hat{\mathbb{Z}}_t$.

We denote the history of instantaneous safety logical states over the past $n$ steps as $\mathbb{Z}_{t-n:t-1} = (\mathbb{Z}_{t-n}, \ldots, \mathbb{Z}_{t-1})$,
where each $\mathbb{Z}_\tau$ is defined in \Eref{equ:define_z}. Under the $n$-th order Markov assumption, only information within this window influences the logical state:
$\hat{\mathbb{Z}}_t \;\perp\!\!\!\perp\; \mathbb{Z}_{1:t-n-1} \mid \mathbb{Z}_{t-n:t-1}$,
\label{eq:markov_independence}
meaning that all evidence for $\hat{\mathbb{Z}}_t$ arises from predicate activations $\{\mathcal{P}_j(e_\tau^{(i)})\}$ within $\tau \in [t-n, t-1]$.

The nMLN consists of a finite set of weighted temporal safety rules
$\Phi = \{(\varphi_k,\, w_k)\}_{k=1}^{K}$, where each $\varphi_k$ is a first-order temporal rule and $w_k \in \mathbb{R}$ is its weight. Intuitively, $\varphi_k$ specifies a pattern of predicates over the past $n$ steps together with a constraint that should be active at time $t$. When grounded on the sequence $(\mathbb{Z}_{t-n:t-1}, \hat{\mathbb{Z}}_t)$, each rule $\varphi_k$ is instantiated on concrete entities, time indices, and constraints in the window and the candidate refined logical state $\hat{\mathbb{Z}}_t$.
Given the recent constraint sequence, the nMLN aggregates evidence from these rules through a log-linear potential:
$\Psi(\mathbb{Z}_{t-n:t-1}, \hat{\mathbb{Z}}_t)
=
\sum_{k=1}^{K} w_k\, f_k(\mathbb{Z}_{t-n:t-1}, \hat{\mathbb{Z}}_t),$
where $f_k(\mathbb{Z}_{t-n:t-1}, \hat{\mathbb{Z}}_t)$ is a function that counts how many groundings of rule $\varphi_k$ are satisfied by the window and the candidate refined logical state. The induced distribution over the logical state is:
\begin{equation}
p(\hat{\mathbb{Z}}_t \mid \mathbb{Z}_{t-n:t-1})
\propto
\exp(\Psi(\mathbb{Z}_{t-n:t-1}, \hat{\mathbb{Z}}_t)),
\end{equation}
which assigns a higher probability to refined logical states
$\hat{\mathbb{Z}}_t$ that are better supported by the temporal safety
evidence encoded in $\Phi$.

Finally, the induced safety state $\hat{\mathbb{Z}}_t$ is fed into the logic-driven revision, enabling \tool to anticipate emerging safety risks and apply context-aware corrections to the base \admlmns.

\subsection{Logic-Driven Action Revision}
\label{sec:rethink}

Given the Markovian-updated refined logical state $\hat{\mathbb{Z}}_t$, \tool checks whether the action proposed by the \admlm is consistent with the inferred active safety constraints. If the proposed action violates any requirement in $\hat{\mathbb{Z}}_t$, \tool triggers a logic-driven action revision step to correct it; otherwise, the original decision is directly accepted. This design is model-agnostic and follows a minimal intervention principle: \tool only revises the output upon constraint violation and leaves safe behaviors unchanged.

We define a violation indicator to test whether the base action $\mathbf{a}_t^{\text{base}}$ in \Eref{equ:action} satisfies the activated constraints inferred from $\hat{\mathbb{Z}}_t$:
\begin{equation}
\delta_t
=
\mathbbm{1}\!\left[\mathbf{a}_t^{\text{base}} \not\models \hat{\mathbb{Z}}_t\right],
\label{eq:violation_indicator}
\end{equation}
where $\delta_t=1$ indicates that $\mathbf{a}_t^{\text{base}}$ violates at least one constraint in $\hat{\mathbb{Z}}_t$ (and $\delta_t=0$ otherwise).

When $\delta_t=1$, \tool does not directly override the action. Instead, it performs action revision by converting the activated $\hat{\mathbb{Z}}_t$ into an auxiliary prompt $\mathbf{p}_t=\mathcal{G}(\hat{\mathbb{Z}}_t)$ and appending it to the current textual input to guide the \admlmns:
\begin{equation}
\label{eq:rethink_prompt}
\tilde{\mathbb{O}}_{t-k:t}
=
\cup_{\tau=t-k}^{t-1} \{(\mathbf{I}_\tau,\, \mathbf{T}_\tau)\}
\ \cup\
\{(\mathbf{I}_{t},\, \mathbf{T}_{t}\oplus \mathbf{p}_t)\},
\end{equation}
where $\oplus$ denotes textual concatenation and $\mathcal{G}(\cdot)$ is a prompt generator that verbalizes violated constraints (\eg, ``Red light detected. Only actions that stop or decelerate are allowed.''). The revised action $\mathbf{a}_t^{\text{re}}$ is then obtained by applying \Eref{equ:action} with $\tilde{\mathbb{O}}_{t-k:t}$ in place of $\mathbb{O}_{t-k:t}$. If $\delta_t=0$, \tool skips action revision and keeps the original output.

\section{Experiments}
\label{sec:experiments}

\subsection{Experimental Setup}
\label{sec:experimental_setup}

\quad \textbf{Models and baselines.} We evaluate three widely used \admlmns: DriveLM \cite{sima2024drivelm}, Dolphins \cite{ma2024dolphins}, and EM-VLM4AD \cite{gopalkrishnanmulti}. For safeguards, we compare \tool (GA) against the closest baseline, SafeAuto (SA) \cite{zhangsafeauto}. We also include two representative safeguards from embodied agents, RoboFactory (RF) \cite{qin2025robofactory} and Code-as-Monitor (CM) \cite{zhou2025code}.

 \textbf{Benchmarks.} We conduct experiments on two benchmarks. For safety-critical scenarios, we use VRU-Accident \cite{kim2025vru}, which collects accident cases targeting \admlmns. For common driving queries and decision-making, we use DriveLM \cite{sima2024drivelm}, a widely used benchmark for \admlm evaluation.

 \textbf{Evaluation Metrics.} We report metrics in two dimensions. For task performance, we use \textit{GPT Score} (GS $\textcolor{red}\uparrow$) and \textit{Qwen Score} (QS $\textcolor{red}\uparrow$), where responses are graded by GPT-4o \cite{achiam2023gpt} and Qwen-3 \cite{yang2025qwen3}, respectively. For safety evaluation on VRU-Accident, we define a benchmark-specific metric, accident rate, to quantify the fraction of cases that result in an accident given the \admlm outputs; it is measured via \textit{GPT Accident Rate} (GAR $\textcolor{blue}\downarrow$) and \textit{Qwen Accident Rate} (QAR $\textcolor{blue}\downarrow$). Each experiment is repeated 10 times, and the average results are reported. \textit{Details of the Accident Rate are in the Appendix.}

 \textbf{Implementation Details.} At each step, the model input includes an observation history of length $k{=}2$. For temporal safety reasoning, we aggregate logical states with an $n$-th order window ($n{=}4$). When a violation is detected, we use GPT-4o \cite{achiam2023gpt} to generate an auxiliary prompt using a fixed verbalization template and re-queries the same \admlm under identical decoding; otherwise, the original action is kept. All experiments are run on a server with a 128-core Intel Xeon 8358 CPU (2.60\,GHz), 1\,TB RAM, and 8$\times$NVIDIA A800 80\,GB PCIe GPUs.

\subsection{Main Results}
\label{sec:Main_Results}

We report the main results of \tool on both safety-critical and common benchmarks.

\subsubsection{Results in Accident-prone Scenarios}
\label{sec:Results_Accident}

We first evaluate \tool on the VRU-Accident benchmark to evaluate performance in accident-prone scenarios. The results are shown in \Tref{tab:main_results_accident}.

\begin{table*}[t]
\Large
\centering
\caption{\textbf{Main results on the VRU-Accident benchmark for different \admlmns, comparing the vanilla model with safeguard methods (RF, CM, SA, GA).} We report the standard deviation ($\sigma$) and the change ($\Delta$). The GA (ours) columns are shaded in gray. Improvements and degradations are highlighted in red and blue, respectively.}
\vspace{-0.05in}
\label{tab:main_results_accident}
\resizebox{\textwidth}{!}{%
\begin{tabular}{cccccccccccccccc}
\toprule[1.2pt]
& \multicolumn{5}{c}{DriveLM} & \multicolumn{5}{c}{Dolphins} & \multicolumn{5}{c}{EM-VLM4AD} \\ \cmidrule(lr){2-6} \cmidrule(lr){7-11} \cmidrule(lr){12-16} 
\multirow{-2}{*}{Method} & Vanilla & +RF & +CM & +SA & \cellcolor{gray!15}+GA & Vanilla & +RF & +CM & +SA & \cellcolor{gray!15}+GA & Vanilla & +RF & +CM & +SA & \cellcolor{gray!15}+GA \\ \midrule
GAR $\textcolor{blue}\downarrow$ & 64.13 & 59.00 & 62.70 & 56.14 & \cellcolor{gray!15}35.61 & 89.33 & 86.49 & 85.92 & 67.17 & \cellcolor{gray!15}32.50 & 53.92 & 52.83 & 53.81 & 50.01 & \cellcolor{gray!15}43.05 \\
$\sigma$ & 0.13 & 0.12 & 0.08 & 0.06 & \cellcolor{gray!15}0.14 & 0.17 & 0.16 & 0.18 & 0.08 & \cellcolor{gray!15}0.20 & 0.12 & 0.13 & 0.05 & 0.05 & \cellcolor{gray!15}0.15 \\
$\Delta$ & / & {\color[HTML]{000070} \textbf{-5.13}} & {\color[HTML]{000070} \textbf{-1.43}} & {\color[HTML]{000070} \textbf{-7.99}} & \cellcolor{gray!15}{\color[HTML]{000070} \textbf{-28.52}} & / & {\color[HTML]{000070} \textbf{-2.84}} & {\color[HTML]{000070} \textbf{-3.41}} & {\color[HTML]{000070} \textbf{-22.16}} & \cellcolor{gray!15}{\color[HTML]{000070} \textbf{-56.83}} & / & {\color[HTML]{000070} \textbf{-1.09}} & {\color[HTML]{000070} \textbf{-0.11}} & {\color[HTML]{000070} \textbf{-3.91}} & \cellcolor{gray!15}{\color[HTML]{000070} \textbf{-10.87}} \\ \midrule
QAR $\textcolor{blue}\downarrow$ & 81.28 & 75.95 & 74.01 & 61.89 & \cellcolor{gray!15}32.80 & 92.35 & 91.84 & 84.71 & 75.55 & \cellcolor{gray!15}41.44 & 56.33 & 55.82 & 55.75 & 52.78 & \cellcolor{gray!15}49.09 \\
$\sigma$ & 0.06 & 0.06 & 0.11 & 0.18 & \cellcolor{gray!15}0.05 & 0.06 & 0.19 & 0.07 & 0.11 & \cellcolor{gray!15}0.19 & 0.06 & 0.12 & 0.08 & 0.07 & \cellcolor{gray!15}0.14 \\
$\Delta$ & / & {\color[HTML]{000070} \textbf{-5.33}} & {\color[HTML]{000070} \textbf{-7.27}} & {\color[HTML]{000070} \textbf{-19.39}} & \cellcolor{gray!15}{\color[HTML]{000070} \textbf{-48.48}} & / & {\color[HTML]{000070} \textbf{-0.51}} & {\color[HTML]{000070} \textbf{-7.64}} & {\color[HTML]{000070} \textbf{-16.80}} & \cellcolor{gray!15}{\color[HTML]{000070} \textbf{-50.91}} & / & {\color[HTML]{000070} \textbf{-0.51}} & {\color[HTML]{000070} \textbf{-0.58}} & {\color[HTML]{000070} \textbf{-3.55}} & \cellcolor{gray!15}{\color[HTML]{000070} \textbf{-7.24}} \\ \midrule
GS $\textcolor{red}\uparrow$ & 42.42 & 43.69 & 43.79 & 49.76 & \cellcolor{gray!15}52.21 & 47.69 & 48.96 & 47.90 & 52.12 & \cellcolor{gray!15}54.72 & 61.47 & 61.73 & 61.51 & 63.96 & \cellcolor{gray!15}65.19 \\
$\sigma$ & 0.18 & 0.20 & 0.05 & 0.20 & \cellcolor{gray!15}0.11 & 0.15 & 0.10 & 0.20 & 0.09 & \cellcolor{gray!15}0.17 & 0.19 & 0.06 & 0.19 & 0.05 & \cellcolor{gray!15}0.13 \\
$\Delta$ & / & {\color[HTML]{700000} \textbf{+1.27}} & {\color[HTML]{700000} \textbf{+1.37}} & {\color[HTML]{700000} \textbf{+7.34}} & \cellcolor{gray!15}{\color[HTML]{700000} \textbf{+9.79}} & / & {\color[HTML]{700000} \textbf{+1.27}} & {\color[HTML]{700000} \textbf{+0.21}} & {\color[HTML]{700000} \textbf{+4.43}} & \cellcolor{gray!15}{\color[HTML]{700000} \textbf{+7.03}} & / & {\color[HTML]{700000} \textbf{+0.26}} & {\color[HTML]{700000} \textbf{+0.04}} & {\color[HTML]{700000} \textbf{+2.49}} & \cellcolor{gray!15}{\color[HTML]{700000} \textbf{+3.72}} \\ \midrule
QS $\textcolor{red}\uparrow$ & 17.99 & 19.21 & 18.23 & 33.83 & \cellcolor{gray!15}42.36 & 10.05 & 14.02 & 10.45 & 35.43 & \cellcolor{gray!15}49.70 & 46.00 & 46.64 & 46.43 & 49.92 & \cellcolor{gray!15}51.37 \\
$\sigma$ & 0.05 & 0.19 & 0.16 & 0.16 & \cellcolor{gray!15}0.14 & 0.17 & 0.14 & 0.20 & 0.09 & \cellcolor{gray!15}0.08 & 0.07 & 0.20 & 0.19 & 0.13 & \cellcolor{gray!15}0.15 \\
$\Delta$ & / & {\color[HTML]{700000} \textbf{+1.22}} & {\color[HTML]{700000} \textbf{+0.24}} & {\color[HTML]{700000} \textbf{+15.84}} & \cellcolor{gray!15}{\color[HTML]{700000} \textbf{+24.37}} & / & {\color[HTML]{700000} \textbf{+3.97}} & {\color[HTML]{700000} \textbf{+0.40}} & {\color[HTML]{700000} \textbf{+25.38}} & \cellcolor{gray!15}{\color[HTML]{700000} \textbf{+39.65}} & / & {\color[HTML]{700000} \textbf{+0.64}} & {\color[HTML]{700000} \textbf{+0.43}} & {\color[HTML]{700000} \textbf{+3.92}} & \cellcolor{gray!15}{\color[HTML]{700000} \textbf{+5.37}} \\ \bottomrule[1.2pt]
\end{tabular}
}
\vspace{-0.05in}
\end{table*}

\ding{182} Without safeguards, all \admlm show accident rates above 50\%, with an average GAR of 68.76\% and QAR of 76.48\%, indicating substantial safety risks and a clear need for effective enhancement methods for \admlmns.

\ding{183} Across all safeguards, \tool yields the largest degradation in accident rate, with average drops of \textbf{32.07\%} (GAR) and \textbf{35.54\%} (QAR), compared to only 5.34\% and 6.84\% on average for others. Meanwhile, all safeguards improve task performance, but \tool delivers the strongest gains, improving GS by \textbf{6.85\%} and QS by \textbf{23.13\%}, whereas the other methods provide smaller average improvements of 2.08\% and 5.78\%. These results indicate that \tool improves task performance while substantially reducing accident risk.

\ding{184} Among all methods, safeguards tailored to \admlm (\ie, \tool and SafeAuto) perform noticeably better than embodied-agent safeguards (\ie, RoboFactory and Code-as-Monitor), highlighting fundamental differences between specific embodied tasks and the AD setting.

\ding{185} We further analyze failure cases. Results show that a large fraction of SafeAuto’s remaining accidents (20.99\%) are triggered by the sudden appearance of traffic participants (e.g., pedestrians). With \toolns, this rate drops to 12.67\%, highlighting the importance of modeling dynamic participants and temporally evolving constraints. \textit{A detailed failure case analysis is provided in \Sref{sec:failure_case_analysis}.}

\subsubsection{Results in Common Scenarios}
\label{sec:results_common}

We further evaluate \tool under common driving settings on the DriveLM benchmark. The results are shown in \Tref{tab:main_results_common}.

\begin{table*}[t]
\Large
\centering
\caption{\textbf{Main results on the DriveLM benchmark for different \admlmns, comparing the vanilla model with safeguard methods (RF, CM, SA, GA).} We report the standard deviation ($\sigma$) and the change ($\Delta$). The GA (ours) columns are shaded in gray. Improvements are highlighted in red.}
\vspace{-0.05in}
\label{tab:main_results_common}
\resizebox{\textwidth}{!}{%
\begin{tabular}{cccccccccccccccc}
\toprule[1.2pt]
& \multicolumn{5}{c}{DriveLM} & \multicolumn{5}{c}{Dolphins} & \multicolumn{5}{c}{EM-VLM4AD} \\ \cmidrule(lr){2-6} \cmidrule(lr){7-11} \cmidrule(lr){12-16} 
\multirow{-2}{*}{Method} & Vanilla & +RF & +CM & +SA & \cellcolor{gray!15}+GA & Vanilla & +RF & +CM & +SA & \cellcolor{gray!15}+GA & Vanilla & +RF & +CM & +SA & \cellcolor{gray!15}+GA \\ \midrule
GS $\textcolor{red}\uparrow$ & 58.39 & 58.78 & 58.60 & 60.08 & \cellcolor{gray!15}61.36 & 44.95 & 45.67 & 45.75 & 47.67 & \cellcolor{gray!15}48.95 & 66.40 & 66.86 & 66.70 & 67.87 & \cellcolor{gray!15}68.42 \\
$\sigma$ & 0.15 & 0.17 & 0.10 & 0.18 & \cellcolor{gray!15}0.07 & 0.06 & 0.09 & 0.12 & 0.08 & \cellcolor{gray!15}0.16 & 0.11 & 0.05 & 0.16 & 0.19 & \cellcolor{gray!15}0.06 \\
$\Delta$ & / & {\color[HTML]{700000} \textbf{+0.39}} & {\color[HTML]{700000} \textbf{+0.21}} & {\color[HTML]{700000} \textbf{+1.69}} & \cellcolor{gray!15}{\color[HTML]{700000} \textbf{+2.97}} & / & {\color[HTML]{700000} \textbf{+0.72}} & {\color[HTML]{700000} \textbf{+0.80}} & {\color[HTML]{700000} \textbf{+2.72}} & \cellcolor{gray!15}{\color[HTML]{700000} \textbf{+4.00}} & / & {\color[HTML]{700000} \textbf{+0.46}} & {\color[HTML]{700000} \textbf{+0.30}} & {\color[HTML]{700000} \textbf{+1.47}} & \cellcolor{gray!15}{\color[HTML]{700000} \textbf{+2.02}} \\ \midrule
QS $\textcolor{red}\uparrow$ & 32.47 & 32.67 & 32.87 & 35.95 & \cellcolor{gray!15}37.51 & 28.43 & 28.56 & 28.62 & 29.40 & \cellcolor{gray!15}30.34 & 41.29 & 41.79 & 41.70 & 42.74 & \cellcolor{gray!15}43.56 \\
$\sigma$ & 0.20 & 0.06 & 0.05 & 0.16 & \cellcolor{gray!15}0.12 & 0.09 & 0.12 & 0.15 & 0.08 & \cellcolor{gray!15}0.09 & 0.07 & 0.19 & 0.08 & 0.17 & \cellcolor{gray!15}0.12 \\
$\Delta$ & / & {\color[HTML]{700000} \textbf{+0.20}} & {\color[HTML]{700000} \textbf{+0.40}} & {\color[HTML]{700000} \textbf{+3.48}} & \cellcolor{gray!15}{\color[HTML]{700000} \textbf{+5.04}} & / & {\color[HTML]{700000} \textbf{+0.13}} & {\color[HTML]{700000} \textbf{+0.19}} & {\color[HTML]{700000} \textbf{+0.97}} & \cellcolor{gray!15}{\color[HTML]{700000} \textbf{+1.91}} & / & {\color[HTML]{700000} \textbf{+0.50}} & {\color[HTML]{700000} \textbf{+0.41}} & {\color[HTML]{700000} \textbf{+1.45}} & \cellcolor{gray!15}{\color[HTML]{700000} \textbf{+2.27}} \\ \bottomrule[1.2pt]
\end{tabular}
}
\end{table*}

\ding{182} All safeguards yield consistent but slight gains on the DriveLM dataset. Across all \admlmns, the average GS and QS are 56.58\% and 34.06\%, respectively. Safeguards provide small average improvements of 1.48\% in GS and 1.41\% in QS, corresponding to relative gains of 2.6\% and 4.1\%.

\ding{183} Among all safeguards, \tool performs best, with average improvements of 3.00\% in GS and 3.07\% in QS. The trend observed in \Sref{sec:Results_Accident} also holds: \admlmsinglens-specific safeguards outperform embodied-agent safeguards, improving GS by 2.48\% vs.\ 0.48\% and QS by 2.52\% vs.\ 0.31\%.

\ding{184} Compared with accident-prone scenarios, the gains in common scenarios are smaller, but safeguards remain consistently beneficial. This suggests that safeguard mechanisms can be applied broadly across driving contexts, rather than being limited to accident-prone settings.

\subsection{Ablation Study}
\label{sec:Ablation}

In this section, we conduct ablation studies on \tool to quantify the impact of each module and key hyperparameters.

\begin{table}[t]
\Huge
\caption{\textbf{Ablation results on the VRU-Accident benchmark using DriveLM.} We report the change ($\Delta$). The \tool row is shaded in gray. Improvements and degradations are highlighted in red and blue, respectively. The best $\Delta$ value in each column is underlined.}
\vspace{-0.05in}
\centering
\label{tab:ablation_modules}
\renewcommand{\arraystretch}{1.2}
\resizebox{\linewidth}{!}{%
\begin{tabular}{@{}ccccccccc@{}}
\toprule[2.0pt]
\multirow{2}{*}{Method} & \multicolumn{4}{c}{Accident Rate} & \multicolumn{4}{c}{Task Performance} \\ \cmidrule[1.2pt](lr){2-5} \cmidrule[1.2pt](lr){6-9} 
 & GAR $\textcolor{blue}\downarrow$ & $\Delta$GAR & QAR $\textcolor{blue}\downarrow$ & $\Delta$QAR & GS $\textcolor{red}\uparrow$ & $\Delta$GS & QS $\textcolor{red}\uparrow$ & $\Delta$QS \\ \midrule[1.2pt]
Base & 64.13 & / & 81.28 & / & 42.42 & / & 17.99 & / \\ \midrule[1.2pt]
\textit{Predicate} & 48.32 & {\color[HTML]{000070} \textbf{-15.81}} & 53.62 & {\color[HTML]{000070} \textbf{-27.66}} & 51.41 & {\color[HTML]{700000} \textbf{+8.99}} & 35.78 & {\color[HTML]{700000} \textbf{+17.79}} \\ 
\textit{Predicate}$^{\dagger}$ & 53.47 & {\color[HTML]{000070} \textbf{-10.66}} & 57.84 & {\color[HTML]{000070} \textbf{-23.44}} & 51.13 & {\color[HTML]{700000} \textbf{+8.71}} & 37.29 & {\color[HTML]{700000} \textbf{+19.30}} \\
\textit{Revision} & 28.74 & \underline{\color[HTML]{000070} \textbf{-35.39}} & 30.17 & \underline{\color[HTML]{000070} \textbf{-51.11}} & 38.16 & {\color[HTML]{000070} \textbf{-4.26}} & 14.56 & {\color[HTML]{000070} \textbf{-3.43}} \\
\textit{Revision}$^{\dagger}$ & 40.26 & {\color[HTML]{000070} \textbf{-23.87}} & 46.52 & {\color[HTML]{000070} \textbf{-34.76}} & 51.24 & {\color[HTML]{700000} \textbf{+8.82}} & 38.16 & {\color[HTML]{700000} \textbf{+20.17}} \\ 
\rowcolor{gray!15} 
\toolns & 35.61 & {\color[HTML]{000070} \textbf{-28.52}} & 32.80 & {\color[HTML]{000070} \textbf{-48.48}} & 52.21 & \underline{\color[HTML]{700000} \textbf{+9.79}} & 42.36 & \underline{\color[HTML]{700000} \textbf{+24.37}} \\ \bottomrule[2.0pt]
\end{tabular}
}
\vspace{-0.05in}
\end{table}

\noindent \textbf{Ablations on Different Modules.}
We evaluate the contribution of each module in \tool on the VRU-Accident benchmark using DriveLM.
We consider ego-centric logic variants: \ding{182} \textit{Predicate} keeps only ego-vehicle and traffic-rule static predicates, and \ding{183} \textit{Predicate}$^{\dagger}$ keeps only predicates relevant to traffic participants.
We also ablate the action revision strategy: \ding{184} \textit{Revision} forces a conservative fallback action upon violation (\eg, \texttt{Stop}), and \ding{185} \textit{Revision}$^{\dagger}$ removes constraint-violating actions and selects the highest-scoring remaining one, isolating constrained selection from model-guided revision. Results in \Tref{tab:ablation_modules} show that \tool outperforms most baselines with comparable module designs. 
A notable exception is \textit{Revision}. By forcing conservative safety actions (\eg, \texttt{Stop} or \texttt{Decelerate}) whenever a violation is detected, it achieves a larger reduction in Accident Rate, but this comes with a clear drop in task performance. In other words, the safety gain is achieved by overly conservative behavior that undermines driving task completion, making it less practical as a safeguard.

\begin{figure}[!t]
    \centering
    \begin{subfigure}{0.24\textwidth}
        \centering
        \includegraphics[width=0.98\linewidth]{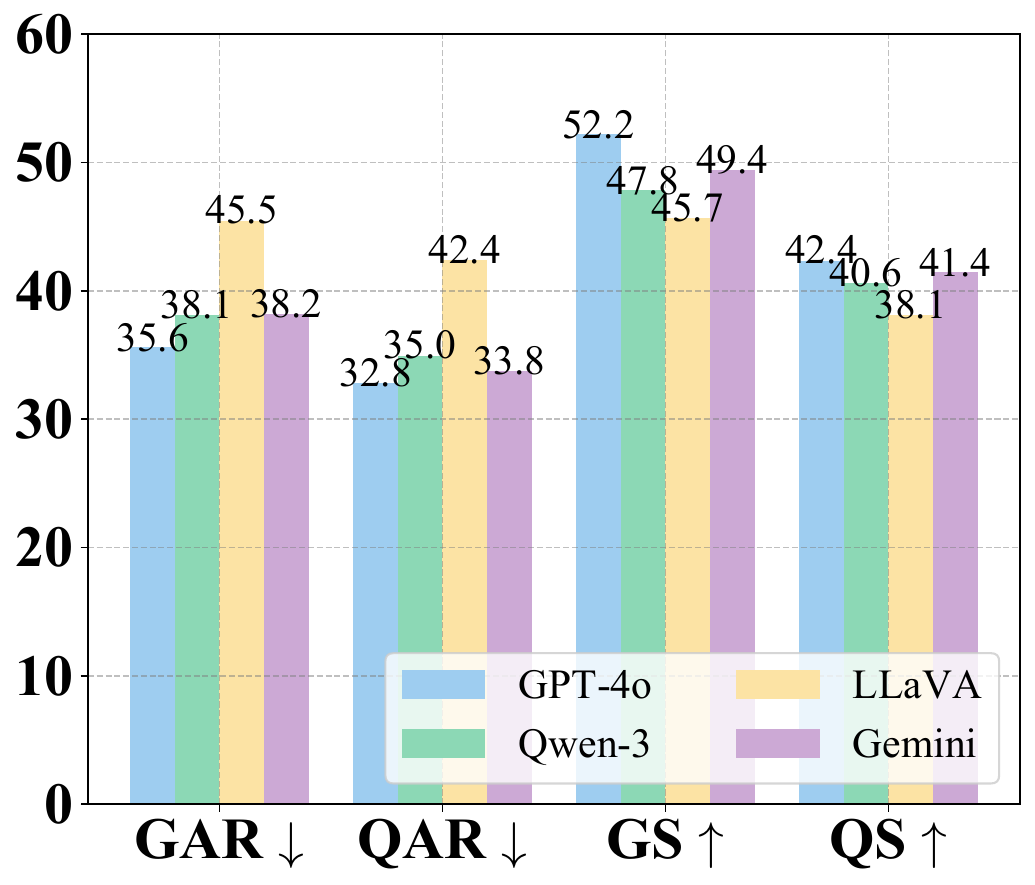}
        \caption{Prompt Generators}
        \label{subfig:different_mllms}
    \end{subfigure}
    \hfill
    \begin{subfigure}{0.23\textwidth}
        \centering
        \includegraphics[width=0.98\textwidth]{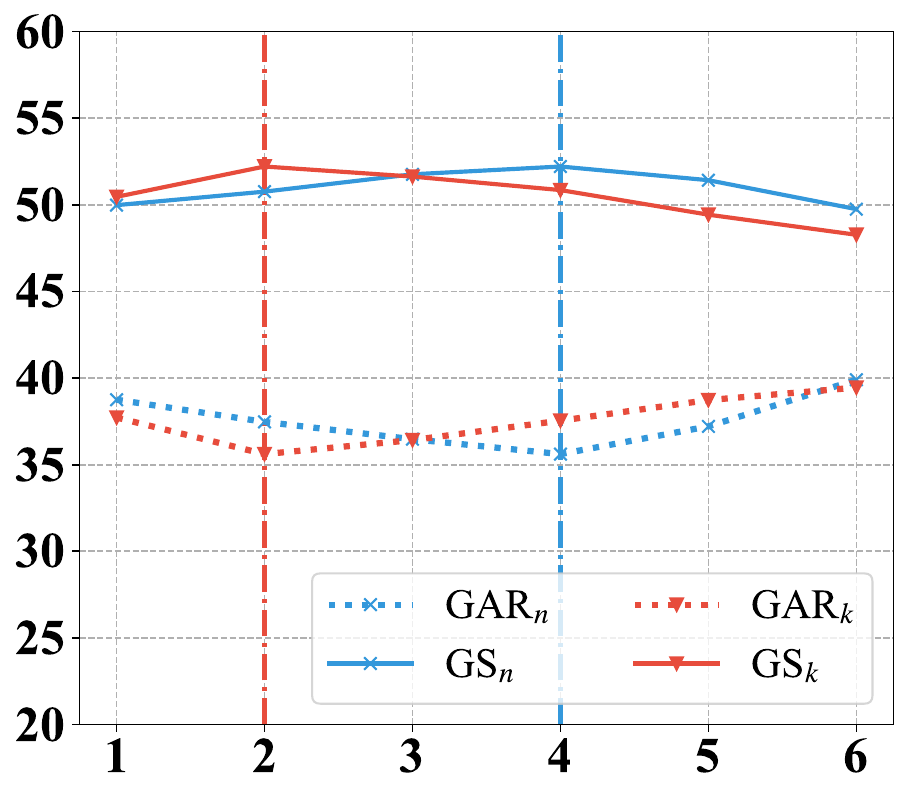}
        \caption{Hyperparameters $n$ and $k$}
        \label{subfig:hyperparameters}
    \end{subfigure}
    \vspace{-0.05in}
    \caption{Ablations on generators and hyperparameters.}
    \vspace{-0.2in}
    \label{fig:ablation}
\end{figure}

\noindent \textbf{Ablations on Prompt Generator $\mathcal{G}(\cdot)$.}
To assess the effect of the prompt generator $\mathcal{G}(\cdot)$ in \Sref{sec:rethink}, we instantiate $\mathcal{G}(\cdot)$ with different MLLMs and compare their performance. We consider four widely used MLLMs: GPT-4o \cite{achiam2023gpt}, Qwen-3 \cite{yang2025qwen3}, LLaVA \cite{liu2023visual_LLaVA}, and Gemini \cite{comanici2025gemini}. Experiments are conducted on VRU-Accident using DriveLM, with results shown in \Fref{subfig:different_mllms}. Overall, GPT-4o achieves the best performance among the tested generators; thus, we use GPT-4o as $\mathcal{G}(\cdot)$ in all experiments unless otherwise specified.

\noindent \textbf{Ablations on Hyperparameter $n$.}
In the nMLN, $n$ controls the Markov order. We vary $n$ and evaluate its impact on both task performance and accident rate on the VRU-Accident benchmark using DriveLM, as shown by the blue curves in \Fref{subfig:hyperparameters}. Overall, $n{=}4$ performs best, achieving the lowest GAR of 35.61\% and the highest GS of 52.21\%, consistent with our main results. When $n{=}1$, the nMLN reduces to a first-order Markov network; it remains effective and still outperforms SafeAuto, which also relies on first-order rule-based checking, suggesting that $n$-th order Markovian Logic Induction provides benefits for \admlmns.

\noindent \textbf{Ablations on Hyperparameter $k$.}
Finally, we study $k$, which controls the observation history length of the \admlmns. We vary $k$ and evaluate its impact on task performance and accident rate on VRU-Accident using DriveLM, as shown by the red curves in \Fref{subfig:hyperparameters}. Overall, $k{=}2$ yields the best results, and we adopt it in the main experiments.
\section{Case Study}
\label{sec:case_study}

Here, we present case studies in both simulation and the physical world to evaluate \toolns.

\subsection{Closed-loop Simulation Evaluation}
\label{sec:closed-loop}

\begin{figure}[t]
    \centering
    \includegraphics[width=0.95\linewidth]{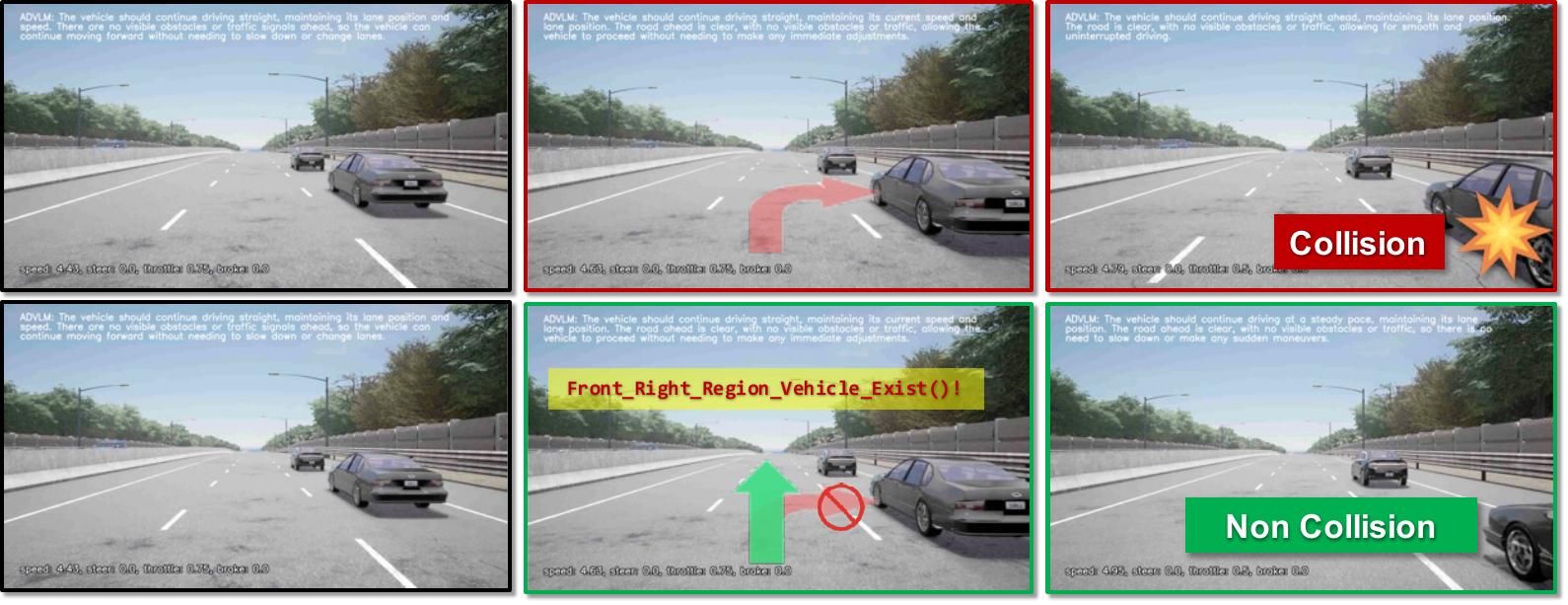}
    \caption{Closed-loop case in CARLA (route scenario 23771). \tool prevents an erroneous right turn and instead triggers a stop, avoiding a collision.}
    \label{fig:simulation}
\end{figure}

Firstly, we evaluate the proposed \tool on Bench2ADVLM \cite{zhang2025bench2advlm} in a closed-loop setting. Concretely, we run CARLA 0.9.15 with DriveLM with LLaMA and compare the vanilla model against the same model equipped with \toolns. Results shown that \tool improves the route success rate from 8.46\% to 12.32\% (+45.60\% relative).
\Fref{fig:simulation} shows a representative route case. With \tool enabled, the model avoids an erroneous right turn and instead goes straight forward, preventing a collision.

\subsection{Human-in-the-loop Physical Evaluation}
\label{sec:human_factors}

\begin{figure}[t]
    \centering
    \includegraphics[width=0.95\linewidth]{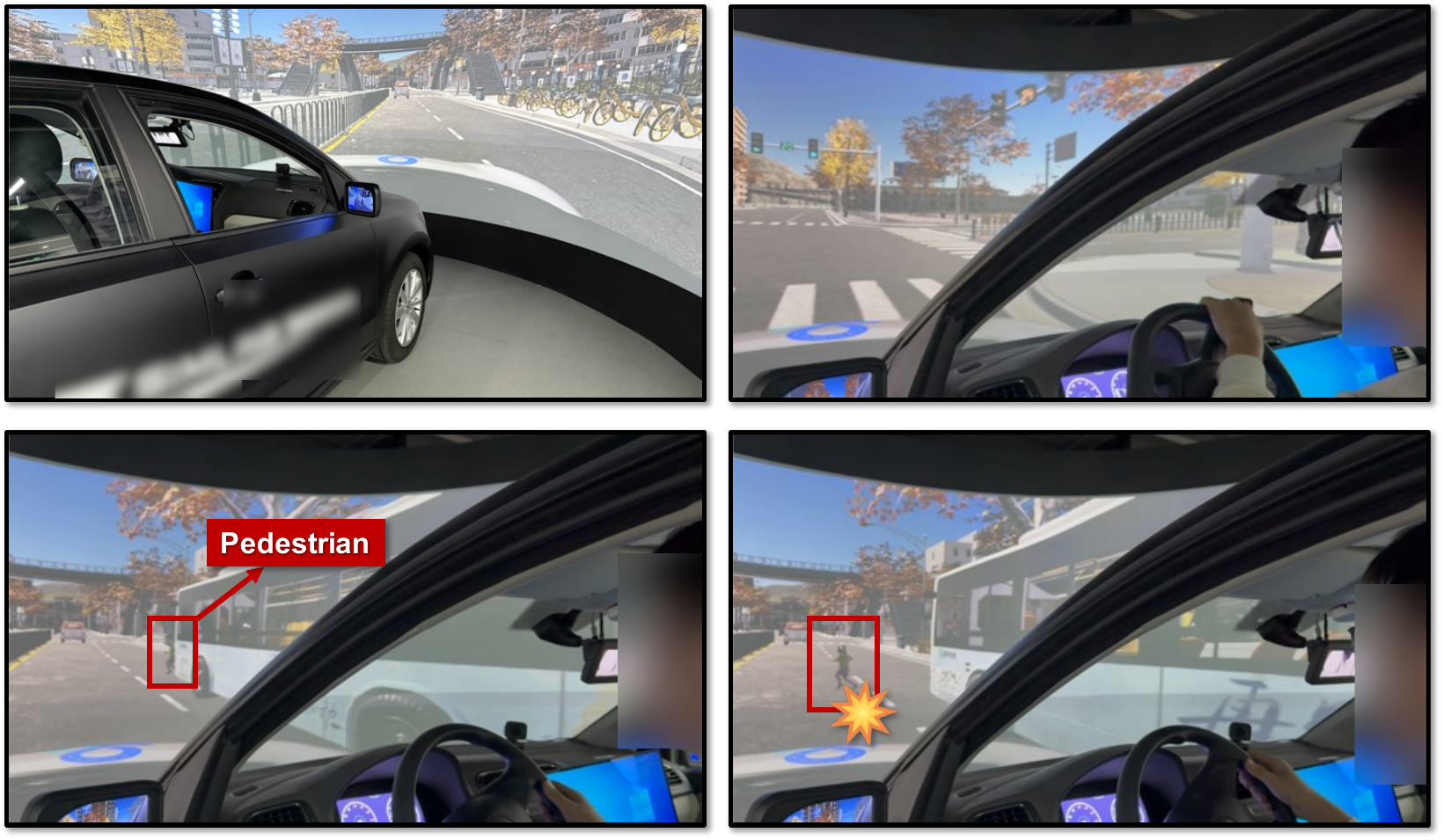}
    \caption{Human-in-the-loop physical-world vehicle evaluation. Experimental setup with a human driver and a representative scenario where a pedestrian suddenly rushes into the ego lane, leading to a potential collision.}
    \vspace{-0.15in}
    \label{fig:Human_experiment}
\end{figure}

Additionally, we conduct a human-in-the-loop study in the lab using a physical-world vehicle. We create a CARLA scenario where a pedestrian suddenly rushes into the ego lane.  An experienced driver operates the vehicle via our vehicle simulator setup while we synchronize and log the driving observations and collision outcomes. We then feed the collected observations to  DriveLM and  DriveLM+\tool to obtain their predicted actions. We compare collision rates across three conditions: \ding{182} human driving, \ding{183} DriveLM, and \ding{184} DriveLM+\toolns. Since the latter two are evaluated offline, we treat a trial as a collision if DriveLM fails to output \texttt{Stop} and \tool does not trigger a stop within a 2.5s window before the annotated collision time \cite{sato2021dirty}. 

Results over 20 trials per condition show collision rates of 35\%, 65\%, and 40\% for human driving, DriveLM, and DriveLM+\toolns. \Fref{fig:Human_experiment} illustrates the setup and a representative collision scenario where a pedestrian suddenly rushes into the lane. The results indicate that \tool substantially reduces collisions and narrows the gap between \admlm and human performance. 
\section{Analysis and Discussion}
\label{sec:discussion}

In this section, we examine \tool more broadly to provide deeper insight.

\subsection{Failure Case Analysis}
\label{sec:failure_case_analysis}

We analyze failure cases in \Sref{sec:Main_Results} and assign each case to one of four mutually exclusive types.
\ding{182} \textit{Ego Decision Error} (EDE): unsafe ego actions without abrupt participant triggers and without explicit rule violations.
\ding{183} \textit{Rule Violation} (RV): violations of traffic rules (\eg, red-light running); RV takes priority when applicable.
\ding{184} \textit{Reactive Participants} (RP): failures triggered by abrupt appearance or motion changes of other participants (\eg, sudden crossings or cut-ins).
\ding{185} \textit{Others} (OT): remaining cases (\eg, predicate grounding errors, instruction ambiguity, or action parsing mismatches).

\begin{table}[t]
\Huge
\caption{\textbf{Failure case analysis on the VRU-Accident benchmark using DriveLM.} We report the number of failures and their proportions. The \tool row is gray, and the lowest value in each column is bold.}
\vspace{-0.05in}
\centering
\label{tab:analysis_failure_case}
\renewcommand{\arraystretch}{1.2}
\resizebox{\linewidth}{!}{%
\begin{tabular}{@{}ccccccccc@{}}
\toprule[2.0pt]
& \multicolumn{2}{c}{EDE} & \multicolumn{2}{c}{RV} & \multicolumn{2}{c}{RP} & \multicolumn{2}{c}{OT} \\ \cmidrule[1.2pt](lr){2-3} \cmidrule[1.2pt](lr){4-5} \cmidrule[1.2pt](lr){6-7} \cmidrule[1.2pt](lr){8-9}
\multirow{-2}{*}{Method} & Amount & Rate (\%) & Amount & Rate (\%) & Amount & Rate (\%) & Amount & Rate (\%) \\ \midrule[1.2pt]
Vanilla & 1302 & \textbf{20.30} & 1395 & 21.76 & 1316 & 20.52 & 2400 & 37.42 \\ 
+RF & 1346 & 22.81 & 1294 & 21.93 & 1351 & 22.90 & 1909 & \textbf{32.36} \\
+CM & 1384 & 22.08 & 1324 & 21.12 & 1502 & 23.96 & 2059 & 32.84 \\
+SA & 1363 & 24.27 & 617 & 10.99 & 1178 & 20.99 & 2456 & 43.75 \\
\rowcolor[HTML]{EFEFEF} +GA & \textbf{1055} & 29.64 & \textbf{369} & \textbf{10.37} & \textbf{451} & \textbf{12.67} & \textbf{1685} & 47.32 \\ \bottomrule[2.0pt]
\end{tabular}
}
\vspace{-0.10in}
\end{table}

As shown in \Tref{tab:analysis_failure_case}, \tool achieves the lowest values across all failure types. SafeAuto mainly reduces RV failures, whereas \tool most clearly outperforms SafeAuto on RP failures. Meanwhile, a substantial portion of EDE and OT failures remains; since participant-triggered hazards or explicit rule violations do not drive these cases, they highlight the need for more effective safeguards for \admlm under such conditions.

\subsection{Defense Against Attacks}
\label{sec:attack}

\begin{table}[t]
\caption{\textbf{Results under attacks on the VRU-Accident benchmark using DriveLM.} We report the change ($\Delta$). The \tool row is shaded in gray. Degradations are highlighted in blue, respectively.}
\vspace{-0.05in}
\centering
\label{tab:discussion_attacks}
\renewcommand{\arraystretch}{0.85}
\resizebox{\linewidth}{!}{%
\begin{tabular}{@{}cccccccc@{}}
\toprule
 & Clean & \multicolumn{2}{c}{CoA} & \multicolumn{2}{c}{ADvLM} & \multicolumn{2}{c}{CAD} \\ \cmidrule(l){2-2} \cmidrule(l){3-4} \cmidrule(l){5-6} \cmidrule(l){7-8}
\multirow{-2}{*}{Method} & GAR & GAR & $\Delta$GAR & GAR & $\Delta$GAR & GAR & $\Delta$GAR \\ \midrule
Vanilla & 64.13 & 83.13 & / & 81.13 & / & 82.13 & / \\ 
+RF & 59.00 & 78.00 & {\color[HTML]{3531FF} \textbf{-5.13}} & 77.00 & {\color[HTML]{3531FF} \textbf{-4.13}} & 77.00 & {\color[HTML]{3531FF} \textbf{-5.13}} \\
+CM & 62.70 & 77.70 & {\color[HTML]{3531FF} \textbf{-5.43}} & 79.70 & {\color[HTML]{3531FF} \textbf{-1.43}} & 77.70 & {\color[HTML]{3531FF} \textbf{-4.43}} \\
+SA & 56.14 & 74.14 & {\color[HTML]{3531FF} \textbf{-8.99}} & 73.14 & {\color[HTML]{3531FF} \textbf{-7.99}} & 70.14 & {\color[HTML]{3531FF} \textbf{-11.99}} \\
\rowcolor[HTML]{EFEFEF} +GA & 35.61 & 50.61 & {\color[HTML]{3531FF} \textbf{-32.52}} & 50.61 & {\color[HTML]{3531FF} \textbf{-30.52}} & 52.61 & {\color[HTML]{3531FF} \textbf{-29.52}} \\ \bottomrule
\end{tabular}
}
\end{table}

To further evaluate the generalization of proposed \toolns, we extend our study to attacks. Specifically, we consider CoA \cite{yang2024chain}, ADvLM \cite{zhang2024visual}, and CAD \cite{wang2025black}, and evaluate DriveLM on the VRU-Accident benchmark. As shown in \Tref{tab:discussion_attacks}, \tool consistently reduces accident rates under diverse attacks (average -23.14\% degradation of GAR), outperforming the other safeguards. Nonetheless, accident rates still increase compared to the non-attack setting even with \tool enabled, highlighting the need for stronger defenses against adversarial attacks.

\subsection{Efficiency Analysis}
\label{sec:efficiency_time}

\begin{table}[t]
\Huge
\caption{\textbf{Efficiency analysis.} We report inference time (seconds) and standard deviation ($\sigma$) over 10 runs. The \tool column is shaded in gray. The lowest time in each row is in bold.}
\vspace{-0.05in}
\centering
\label{tab:discussion_efficiency}
\renewcommand{\arraystretch}{1.1}
\resizebox{\linewidth}{!}{%
\begin{tabular}{cccccccc
>{\columncolor[HTML]{EFEFEF}}c 
>{\columncolor[HTML]{EFEFEF}}c}
\toprule[2.0pt]
 & Vanilla & \multicolumn{2}{c}{+RF} & \multicolumn{2}{c}{+CM} & \multicolumn{2}{c}{+SA} & \multicolumn{2}{c}{\cellcolor[HTML]{EFEFEF}+GA} \\ \cmidrule[1.2pt](l){2-2} \cmidrule[1.2pt](l){3-4} \cmidrule[1.2pt](l){5-6} \cmidrule[1.2pt](l){7-8} \cmidrule[1.2pt](l){9-10}
\multirow{-2}{*}{Method} & Time & Time & $\sigma$ & Time & $\sigma$ & Time & $\sigma$ & Time & $\sigma$ \\ \midrule[1.2pt]
DriveLM & 1.36 & 1.89 & 0.16 & \textbf{1.76} & 0.25 & 1.79 & 0.24 & 1.77 & 0.25 \\
Dolphins & 2.75 & 4.23 & 0.48 & 3.88 & 0.67 & \textbf{3.49} & 0.59 & 3.54 & 0.56 \\ \bottomrule[2.0pt]
\end{tabular}
}
\end{table}

We analyze the runtime efficiency on DriveLM and Dolphins on the VRU-Accident using the server mentioned in \Sref{sec:experimental_setup}. Since a safeguard is only triggered when violations are detected, we report the average latency over 10 runs in \Tref{tab:discussion_efficiency}. Overall, all safeguards introduce only a small overhead compared to the vanilla \admlm inference time. \tool achieves near-best efficiency, with average latency 1.77 for DriveLM and 3.54 for Dolphins, only slightly slower than the fastest baseline.

\section{Conclusion and Future Work}
\label{sec:conclusion}

We presented \toolns, a model-agnostic safeguard for \admlm that formulates AD safety as an evolving Markovian logical state. It encodes Neuro-Symbolic Logic Formalization and updates the state with an $n$-th order Markovian Logic Induction, then performs Logic-Driven Action Revision to refine unsafe actions without modifying the \admlmns. Extensive experiments indicate \toolns's effectiveness and applicability.

\noindent \textbf{Future Work.} \ding{182} Improve grounding and parsing (\eg, entity tracking, predicate extraction). \ding{183} Expand predicates to cover ego decision error.


\section*{Limitations}
\label{sec:limiations}

While the \tool shows promising results, several limitations remain.
\ding{182} Its effectiveness relies on accurate entity, predicate, and action grounding; perception or extraction errors may propagate and cause unresolved failures.
\ding{183} The current predicate and rule set does not fully capture all ego-centric decision errors or open-ended driving intents.
\ding{184} The revision process incurs additional computational overhead and depends on external generators.

\section*{Ethics Statement}
\label{sec:Ethics}

This paper proposes \toolns, a safeguard that reduces unsafe decisions made by autonomous-driving multimodal large language models through Markovian Safety Logic, rather than enhancing capabilities that could be misused. The study relies on widely used autonomous driving benchmarks and closed-loop simulation, and does not involve the collection of sensitive personal data or any form of profiling. A small human-in-the-loop evaluation is conducted in a controlled laboratory setting with an experienced driver operating the vehicle; model behavior is assessed offline using logged observations, keeping participant risk minimal and avoiding exposure from real-world deployment.

\bibliography{custom}



\end{document}